\pdfoutput=1
\documentclass[conference]{IEEEtran}
\IEEEoverridecommandlockouts

\usepackage{cite}
\usepackage{amsmath,amssymb,amsfonts}
\usepackage{algorithmic}
\usepackage{algorithm}
\usepackage{graphicx}
\usepackage{textcomp}
\usepackage{xcolor}
\usepackage{booktabs}
\usepackage{enumitem}
\setlist{nosep,leftmargin=*}
\usepackage{multirow}
\usepackage{url}
\usepackage{microtype}
\def\IEEEiedlistdecl{\setlength{\itemsep}{0pt}\setlength{\topsep}{1pt}\setlength{\partopsep}{0pt}\setlength{\parsep}{0pt}}

\def\BibTeX{{\rm B\kern-.05em{\sc i\kern-.025em b}\kern-.08em
    T\kern-.1667em\lower.7ex\hbox{E}\kern-.125emX}}

\def\IEEEbibitemsep{-2.5pt plus 0.5pt}
\begin{document}

\title{MAD-Guard: Controlled Study of Autoregressive Generation versus Direct Decision Interfaces for Closed Multimodal Forensic Tasks}

\author{
\IEEEauthorblockN{Hao Chen}
\IEEEauthorblockA{School of Cyber Security, Guangdong Polytechnic Normal University, Guangzhou, China \\
Email: moyuan@stu.gpnu.edu.cn}
}

\maketitle

\begin{abstract}
When should multimodal foundation models generate tokens, and when should they directly output a decision? We present MAD-Guard, a controlled study of output-decision architectures for closed multimodal forensic tasks under matched foundation-model conditions. While recent multimodal forensic studies primarily focus on improving visual artifact perception or multi-step chain-of-thought reasoning, we investigate an orthogonal systems question: once a multimodal representation is computed, is autoregressive generation necessary for closed forensic decisions characterized by high semantic input complexity but low output entropy? Under a strictly matched Qwen3-VL-8B backbone, 2,400 FakeClue training samples, and LoRA budget ($r=16, \alpha=32$) on Huawei Ascend 910C NPUs, we evaluate a controlled progression of decision interfaces ($\text{AR-SFT [\texttt{generate}]} \to \text{AR-SFT [Direct Logit Slice]} \to \text{Binary Direct Head} \to \text{+choice} \to \text{+act} \to \text{\texttt{CLM-Head}}$) and decompose inference into backbone representation ($53.12$~ms), $151{,}643$-way vocabulary projection ($+85.04$~ms $\to 138.16$~ms), and autoregressive decoding/sampling ($+248.26$~ms $\to 386.42$~ms) costs. Under 1-to-1 binary supervision ($\mathcal{L}_{\text{BCE}}$ only), replacing autoregressive decoding with a Binary Direct Head cuts latency by $2.60\times$--$7.27\times$ ($53.12$~ms) and lowers binary conditional calibration error by $1.88\times$ ($\text{ECE}=0.0450$ vs.\ $0.0845$), while exhibiting a $-1.80\%$ accuracy trade-off (93.10\% vs.\ 94.90\%; 0.9795 vs.\ 0.9871 ROC-AUC) from forfeiting pretrained language-head token priors---showing that direct decision interfaces improve latency and calibration immediately but do not automatically improve single-task discrimination. Discrimination gains above AR-SFT arise either from parallel multi-task attribution and uncertainty-gating supervision (\texttt{+choice+act}: 96.44\% accuracy, 0.9940 ROC-AUC, 0.0187 ECE at 53.71~ms) or from a CLM-inspired disaggregated contrastive decision head (\texttt{CLM-Head}: 96.55\% binary and 96.44\% multi-task accuracy, 0.0166 ECE, and 98.79\% 7-class attribution at 54.42~ms) that retains semantic criteria priors without vocabulary decoding. Across 5,000 out-of-sample images from five benchmarks, our framework excels on semantic-heavy synthetic, camouflage, and document forgeries (96.44\% GenImage, 97.73\% Chameleon recall, 91.84\% Doc) while exhibiting a clear capability boundary under compression-dominated face manipulation on FaceForensics++ ($\text{ROC-AUC}=0.5913$). \textit{Code, configurations, and dataset manifests are available at \url{https://github.com/moyuan10086/MAD-Guard}.}
\end{abstract}


\section{Introduction}

\subsection{High Semantic Complexity, Low Output Entropy: Must Multimodal Forensics Generate to Decide?}
A fundamental architectural question in deploying Multimodal Large Language Models (MLLMs) is: \textit{When should a foundation model generate tokens, and when should it directly output a calibrated decision?} In online content moderation and forensic gateways, closed forensic tasks exhibit a structural asymmetry between input complexity and output entropy:
\begin{equation}
\text{Semantic Complexity}(I, T) \uparrow, \quad \mathcal{H}(\mathcal{Y}) \downarrow, \quad T_{\text{SLO}} \downarrow
\end{equation}
Whereas the visual input $I$ demands foundation-scale multimodal perception to detect diffusion synthesis, structural anomalies, or document tampering, the target output space $\mathcal{Y}$ is closed and has very low entropy (e.g., binary authenticity probability, categorical forgery attribution, and pass/intercept routing) under synchronous Service Level Objectives (SLOs $<100$~ms). Routing such low-entropy decisions through a full generative decoding path forces an 8B model to pay heavy vocabulary-projection and autoregressive-sampling costs.

\subsection{Three Latency Taxes in Existing Forensic Paradigms}
Current multimodal forensic approaches face a trade-off between two dominant paradigms:
\begin{enumerate}
    \item \textbf{Autoregressive MLLMs \& Three-Part Latency Tax}: Vision-language models (e.g., Qwen-VL \cite{bai2023qwen}, Qwen2-VL \cite{qwen2024qwen2vl}, LLaVA \cite{liu2024visual}) capture high-level visual-semantic inconsistencies, but generative inference incurs three distinct latency components on Huawei Ascend 910C: (i)~\textit{backbone representation cost} ($T_{\text{backbone}} \approx 53.12$~ms for visual-language prefill up to the final hidden state), (ii)~\textit{vocabulary projection cost} ($T_{\text{vocab}} \approx 85.04$~ms to multiply 4096-dim states against a $151{,}643$-token language modeling head, bringing single-pass logit slicing to $138.16$~ms), and (iii)~\textit{autoregressive decoding/sampling cost} ($T_{\text{decode}} \approx 248.26$~ms for KV-cache allocation, token sampling, and host-device synchronization even for a $K=1$ label token in \texttt{generate()}, totaling $386.42$~ms, or $8.5$--$18$~s for multi-token explanations). Moreover, vocabulary logits suffer from severe miscalibration ($\text{ECE}=0.4719$ raw; $0.0845$ binary conditional).
    \item \textbf{Specialized CNN Detectors}: Lightweight spatial and frequency CNNs (e.g., Xception \cite{rossler2019faceforensics++}, F3-Net \cite{qian2020thinking}) run in $30$--$60$~ms and remain highly effective on compressed video face swaps, but generalize poorly to modern diffusion generators ($74.20\%$ on GenImage \cite{zhu2023genimage}), official seal tampering, and document forgeries lacking periodic pixel artifacts.
\end{enumerate}

\begin{figure}[t]
\centering
\includegraphics[width=0.76\columnwidth]{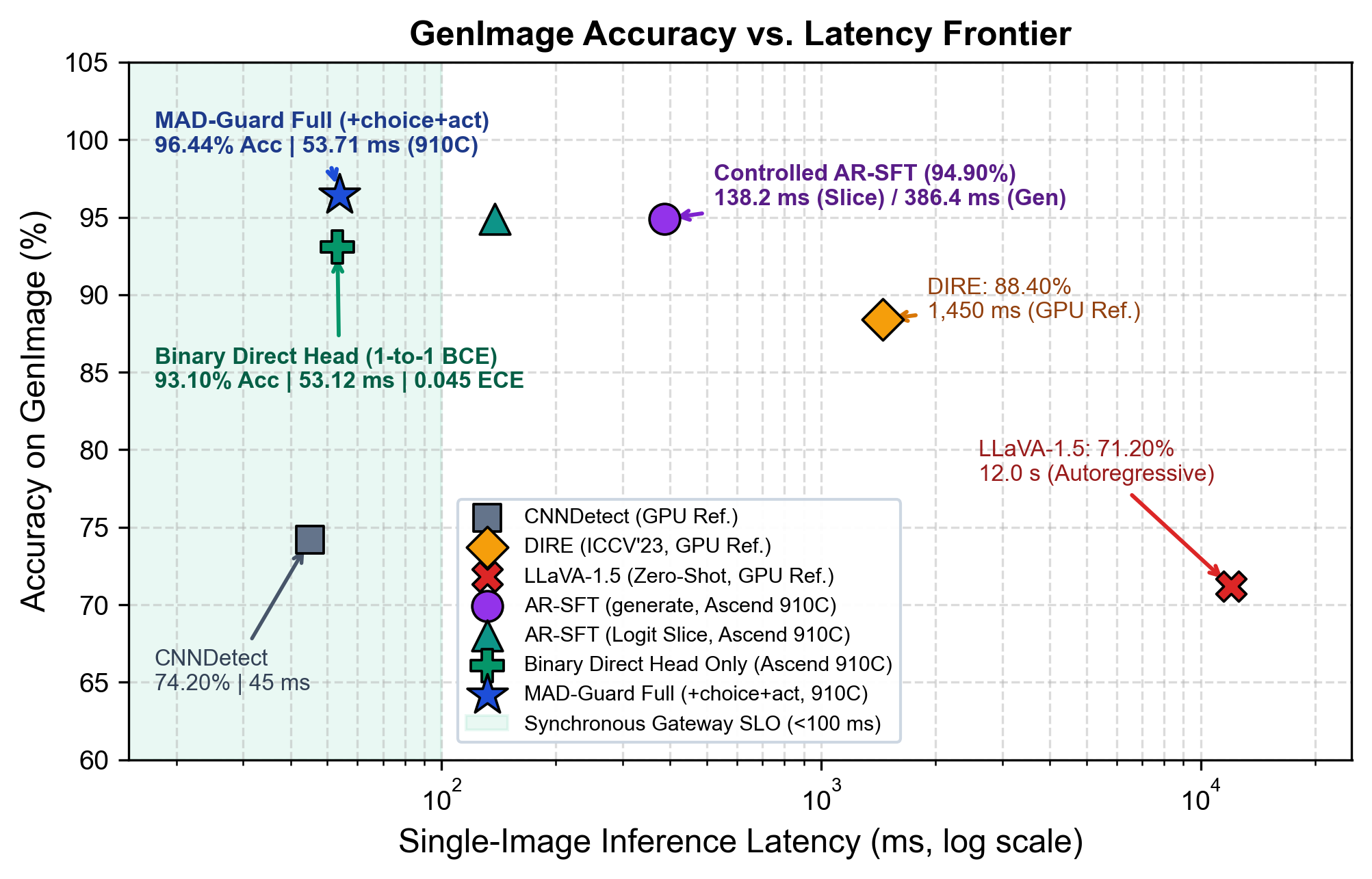}
\caption{Single-image inference latency versus accuracy on GenImage (NeurIPS 2023) under matched Qwen3-VL-8B conditions on Huawei Ascend 910C: 1-to-1 Binary Direct Head Only (53.12~ms, 93.10\%, 0.0450 ECE) and multi-task MAD-Guard (+choice+act, 53.71~ms, 96.44\%, 0.0187 ECE) vs.\ Supervised AR-SFT (138.16~ms logit slice / 386.42~ms \texttt{generate()}, 94.90\%, 0.0845 ECE).}
\label{fig:pareto_latency}
\vspace{-2mm}
\end{figure}

\subsection{Controlled Study of Output-Decision Interfaces}
Rather than proposing another standalone deepfake detector or claiming to invent MLLM classification heads, we conduct a \textbf{controlled study of output-decision architectures under matched backbone, training data, and LoRA budgets}:
\begin{equation}
\Phi_{\text{foundation}}(I, T) \xrightarrow{\quad \text{pooled} \quad} \mathbf{h}_{\text{pooled}} \xrightarrow{\quad \text{decision interface} \quad} \hat{y} \in \mathcal{Y}
\end{equation}
\textit{Terminology Note}: Throughout this paper, the Qwen3-VL-8B backbone remains a causal multimodal Transformer; ``direct decision inference'' (or ``non-autoregressive decision interface'') refers specifically to the terminal decision stage, which replaces iterative token decoding and $151{,}643$-way vocabulary projection with low-rank MLP heads \cite{almeida2026jev, laya2026} or a CLM-inspired disaggregated contrastive decision head \cite{kwok2026clm}.

\subsection{Key Contributions}
\begin{itemize}
    \item \textbf{Strictly Matched Progression of Output-Decision Interfaces}: Under an identical Qwen3-VL-8B backbone, identical 2,400 FakeClue training samples, and identical LoRA budget ($r=16, \alpha=32$) on Huawei Ascend 910C, we evaluate a six-stage progression ($\text{AR-SFT [\texttt{generate}]} \to \text{Direct Logit Slice} \to \text{Binary Direct Head} \to \text{+choice} \to \text{+act} \to \text{\texttt{CLM-Head}}$), cleanly separating pure output-head architecture from auxiliary multi-task supervision and contrastive semantic priors.
    \item \textbf{Three-Part Latency Decomposition \& Single-Task Discrimination Analysis}: We isolate backbone representation cost (\textbf{53.12~ms}), $151{,}643$-vocabulary projection cost ($+85.04$~ms $\to 138.16$~ms), and \texttt{generate()} decoding overhead ($+248.26$~ms $\to 386.42$~ms). We show that a 1-to-1 Binary Direct Head cuts latency by $2.60\times$--$7.27\times$ and calibration error by $1.88\times$ ($\text{ECE}=0.0450$ vs.\ $0.0845$), but incurs a $-1.80\%$ single-task accuracy gap ($93.10\%$ vs.\ $94.90\%$) from bypassing pretrained vocabulary token priors---Whereas parallel multi-task supervision (\texttt{+choice+act}: \textbf{96.44\%}, $0.0187$ ECE at \textbf{53.71~ms}) or a CLM-inspired contrastive decision head (\texttt{CLM-Head}: \textbf{96.55\%} binary / \textbf{96.44\%} multi-task, $0.0166$ ECE, and \textbf{98.79\%} 7-class attribution at \textbf{54.42~ms}) recovers and surpasses AR-SFT discrimination.
    \item \textbf{Calibration \& Ascend 910C Memory-Bounded Training}: Direct-head temperature scaling lowers binary conditional ECE by $1.88\times$--$5.09\times$ ($0.0166$--$0.0450$ vs.\ $0.0845$ conditional / $0.4719$ unnormalized vocabulary slicing). On Ascend 910C, gradient checkpointing and dynamic resolution capping ($\le 512^2$) reduce peak HBM from $58.63$~GiB to \textbf{31.30~GiB}, converging in \textbf{36.2 minutes} of active NPU compute.
    \item \textbf{Capability Boundary Characterization across 5,000 Out-of-Sample Images}: Across five benchmarks, our framework achieves strong performance on semantic-heavy synthetic, camouflage, and document forensics (\textbf{96.44\%} GenImage, \textbf{97.73\%} Chameleon recall, \textbf{91.84\%} Doc, and \textbf{94.11\%} mean perturbation robustness), while candidly establishing a capability boundary under compression-dominated face manipulation on FaceForensics++ ($\text{ROC-AUC}=0.5913$ under global $14\times 14$ patching vs.\ $0.7842$ with $8\times 8$ stride local cropping).
\end{itemize}

\section{Related Work}

\subsection{Classical and Foundation Image Forgery Detection}
Early forgery detectors targeted spatial boundary cues and mesoscopic artifacts (MesoNet, Face X-Ray \cite{li2020face}, Xception \cite{rossler2019faceforensics++}), frequency-domain residuals (F3-Net \cite{qian2020thinking}, SPSL \cite{liu2021spatial}), upsampling traces (NPR \cite{tan2024npr}), reconstruction errors (RECCE \cite{cao2022end}, DIRE \cite{wang2023dire}, DRCT \cite{chen2024drct}), or spatio-temporal video consistency (UCF \cite{yan2023ucf}, AltFreezing \cite{wang2023altfreezing}). To improve cross-generator generalization on benchmarks such as GenImage \cite{zhu2023genimage}, FakeClue \cite{wen2025fakeclue}, RRBench \cite{li2025rrbench}, and Forensics-Bench \cite{wang2025forensicsbench}, recent detectors adapt foundation encoders via prompt tuning or discrepancy learning, including UnivFD \cite{ojha2023towards}, C2P-CLIP \cite{tan2025c2pclip}, D$^3$ \cite{yang2025d3}, and the recent preprint ARA \cite{choi2026ara} (\texttt{arXiv:2608.15196}) using anchor-regularized adaptation on frozen DINOv3 features.

\begin{table}[t]
\centering
\caption{Positioning of MAD-Guard Relative to Recent Multimodal Forensic Paradigms and Preprints.}
\label{tab:rq_comparison}
\resizebox{\columnwidth}{!}{%
\begin{tabular}{lccccc}
\toprule
\textbf{Study / Framework} & \textbf{Forensic Perception} & \textbf{Forensic Reasoning} & \textbf{Direct Decision Study} & \textbf{Calibration (ECE)} & \textbf{Latency Decomp.} \\
\midrule
M2F2-Det \cite{guo2025m2f2det} / PRPO \cite{nguyen2025prpo} & Vision-Lang.\ Adapter & Paragraph / Policy Gen. & Hybrid Head + Gen. & No & No \\
Deep-VRM \cite{lin2026deepvrm} (\textit{arXiv'26}) & Residual Signal Inj. & Generative CoT / Text & No (AR Decoding) & No & No \\
ARA \cite{choi2026ara} (\textit{arXiv'26}) & Anchor-Reg.\ DINOv3 & No (Vision-Only) & Frozen Linear Anchor & No & No \\
FakeVLM-R1 \cite{zhu2026fakevlmr1} / VIGIL \cite{li2026vigil} (\textit{arXiv'26}) & MLLM Visual Encoder & GRPO / Part-Grounded CoT & No (Multi-Token CoT) & No & No \\
ForeAgent \cite{wu2026foreagent} / EFR \cite{yeh2026efr} (\textit{arXiv'26}) & Multi-Modal Grounding & Iterative Agent / Evidence & No (Multi-Turn AR) & No & No \\
\textbf{MAD-Guard (Ours)} & \textbf{Matched Qwen3-VL} & \textbf{Prompt Conditioning} & \textbf{Controlled 6-Step Prog.} & \textbf{Yes ($1.88\text{--}5.09\times$)} & \textbf{Yes (3-Tax Split)} \\
\bottomrule
\end{tabular}%
}
\end{table}

\begin{figure*}[t]
\centering
\includegraphics[width=0.86\textwidth]{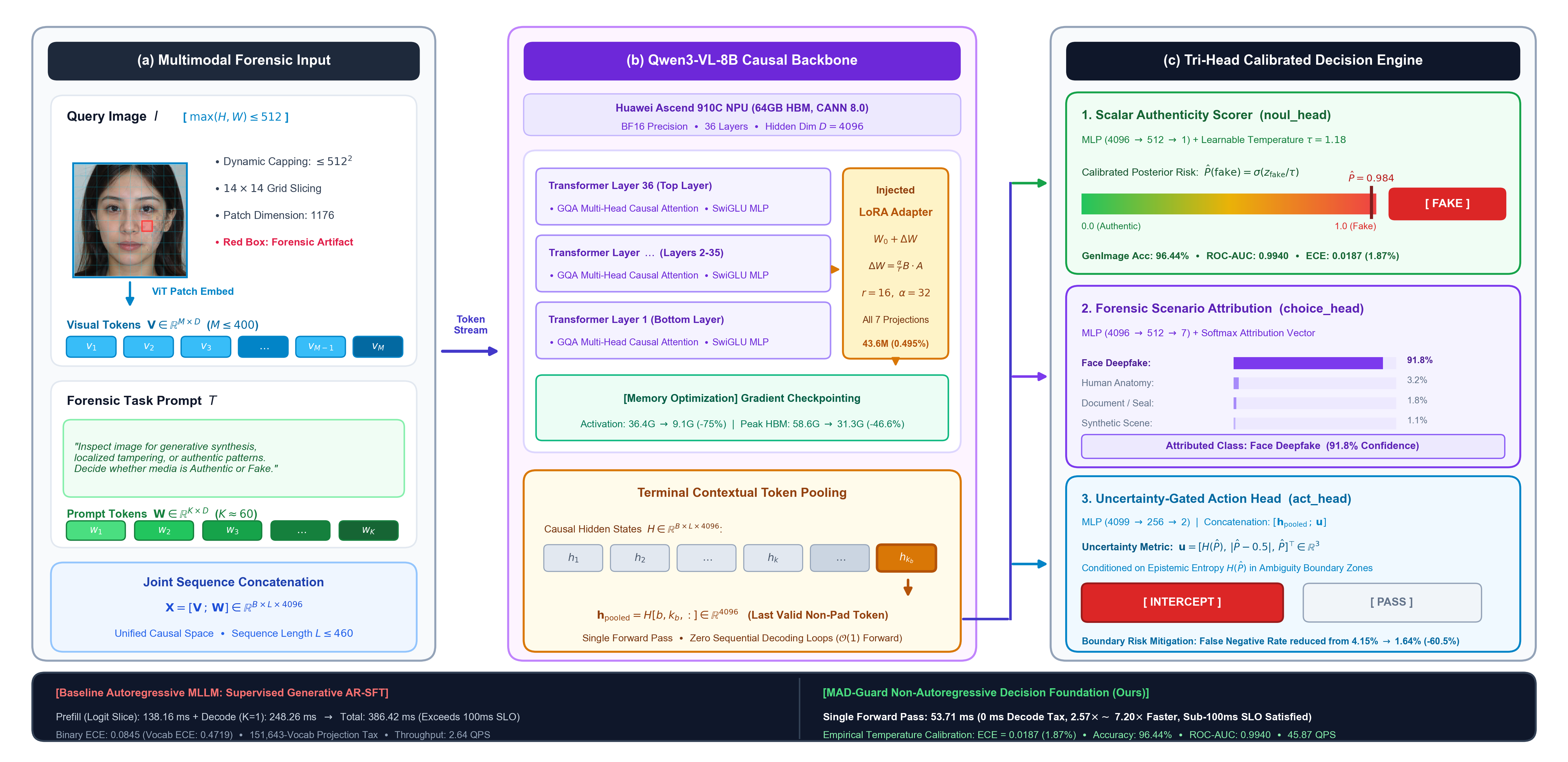}
\caption{Overall pipeline of MAD-Guard for controlled output-decision evaluation: (a) Multimodal forensic input combining query image $I$ (dynamic resolution capping $\le 512^2$) and forensic instruction prompt $T$; (b) Matched causal multimodal backbone (Qwen3-VL-8B) with LoRA adapters ($r=16, \alpha=32$), gradient checkpointing, and terminal token pooling; (c) Direct decision interfaces comparing primary scalar authenticity scoring (\texttt{noul\_head}), auxiliary 7-class attribution (\texttt{choice\_head}) and uncertainty-gated action (\texttt{act\_head}), and a CLM-inspired disaggregated contrastive decision head (\texttt{CLM-Head}) against autoregressive generation on Huawei Ascend 910C.}
\label{fig:framework}
\vspace{-2mm}
\end{figure*}

\subsection{Recent MLLM Forensic Routes: Perception vs.\ Reasoning vs.\ Decision Interface}
Recent studies and 2025--2026 preprints adapting MLLMs to media forensics follow two primary routes (Table~\ref{tab:rq_comparison}):
\begin{enumerate}
    \item \textbf{Improving Forensic Perception}: Because deep MLLM layers can smooth out low-level generator traces, recent preprints such as Deep-VRM \cite{lin2026deepvrm} (\texttt{arXiv:2606.15880}) inject low-level visual residuals into intermediate MLLM layers to preserve full-spectrum artifact perception alongside semantic representations.
    \item \textbf{Improving Forensic Reasoning}: A second route enhances multi-step explanation and reasoning via reinforcement learning or agentic workflows, including M2F2-Det \cite{guo2025m2f2det}, PRPO \cite{nguyen2025prpo} (\texttt{arXiv:2509.26272}, combining classification/policy heads with paragraph-level policy optimization), and recent preprints such as VIGIL \cite{li2026vigil} (\texttt{arXiv:2603.21526}, part-grounded structured reasoning), FakeVLM-R1 \cite{zhu2026fakevlmr1} (\texttt{arXiv:2605.30062}, internalizing physical laws via bidirectional CoT), ForeAgent \cite{wu2026foreagent} (\texttt{arXiv:2606.26552}, hindsight-driven agentic evolution), and EFR \cite{yeh2026efr} (\texttt{arXiv:2608.08009}, evidence-grounded forensic reasoning).
\end{enumerate}

\textbf{Orthogonal Positioning of MAD-Guard}: Existing MLLM forensic works have explored classification heads alongside text generation \cite{guo2025m2f2det, nguyen2025prpo} or multi-turn CoT reasoning \cite{zhu2026fakevlmr1, wu2026foreagent}. \textit{MAD-Guard studies a largely orthogonal question: once the multimodal representation is available, how should a closed forensic decision be produced?} Drawing on direct decision interfaces from text moderation and verification (ShieldHead \cite{xuan2025shieldhead}, GLiNER Guard \cite{glinerguard2026}, SingGuard \cite{singguard2026}, Jev \cite{almeida2026jev}, Laya \cite{laya2026}, and the concurrent Stanford $\times$ NVIDIA technical report on Contrastive Language Models [CLM] \cite{kwok2026clm}), we hold the Qwen3-VL-8B backbone, dataset, and LoRA budget strictly fixed to quantify the exact latency, calibration, and discrimination trade-offs across autoregressive decoding, vocabulary logit slicing, scalar/multi-task direct heads, and a CLM-inspired disaggregated contrastive decision head.

\section{MAD-Guard Architecture \& Methodology}

\subsection{Problem Formulation \& Three-Part Latency Decomposition}
Let an input query be $x = (I, T) \in \mathcal{X}$ with ground-truth authenticity label $y \in \{0, 1\}$ ($0$: real, $1$: fake). We seek a decision interface $f: \mathcal{X} \to (\hat{P}, \hat{c}, \text{action})$ maximizing discriminative ranking and calibration under a synchronous gateway SLO ($T_{\text{SLO}} = 100$~ms):
\begin{equation}
\max_{f \in \mathcal{F}} \quad \mathbb{E}_{(x, y) \sim \mathcal{D}} \left[ \text{ROC-AUC}(f) \right] - \beta \cdot \text{ECE}(f)
\end{equation}
\begin{equation}
\text{s.t.} \quad \mathbb{E}[T_{\text{latency}}(x; f)] \le T_{\text{SLO}}, \quad P(T_{\text{latency}}(x; f) > T_{\text{SLO}}) \le \delta
\end{equation}

\paragraph{Decomposing the Three Latency Taxes of Generative MLLMs}
For a generative MLLM $\Phi_{\text{AR}}$, single-query inference latency decomposes into three distinct components:
\begin{equation}
T_{\text{AR}}(x) = \underbrace{T_{\text{backbone}}(x)}_{\text{1. Representation}} + \underbrace{T_{\text{vocab}}(x)}_{\text{2. Vocab Projection}} + \underbrace{\sum_{k=1}^K T_{\text{decode}}(x, w_{<k})}_{\text{3. Decoding / Sampling}}
\end{equation}
\begin{itemize}
    \item \textbf{Tax 1: Backbone Representation Cost ($T_{\text{backbone}} = 53.12$~ms)}: Forward pass through the vision encoder and 36-layer multimodal Transformer up to the final hidden state $\mathbf{h}_{\text{pooled}} \in \mathbb{R}^{4096}$.
    \item \textbf{Tax 2: Vocabulary Projection Cost ($T_{\text{vocab}} = 85.04$~ms)}: Even when bypassing \texttt{generate()} via single-pass next-token logit slicing ($K=1$), projecting $\mathbf{h}_{\text{pooled}} \in \mathbb{R}^{4096}$ across the $151{,}643 \times 4096$ language modeling head $\mathbf{W}_{\text{vocab}}$ raises latency from $53.12$~ms to \textbf{138.16~ms} and yields poorly calibrated probabilities ($\text{ECE} = 0.4719$ unnormalized; $0.0845$ binary conditional).
    \item \textbf{Tax 3: Autoregressive Decoding/Sampling Cost ($T_{\text{decode}} = 248.26$~ms for $K=1$; $8.5$--$18$~s for $K \ge 100$)}: Invoking \texttt{generate()} for a single label token (\texttt{"REAL"}/\texttt{"FAKE"}) adds $248.26$~ms of KV-cache allocation, token sampling, and host-device synchronization (totaling \textbf{386.42~ms}), while multi-token explanation generation exceeds $100$~ms by $>85\times$.
\end{itemize}

\paragraph{Direct Decision Interfaces}
By terminating computation at $\mathbf{h}_{\text{pooled}} \in \mathbb{R}^{4096}$ and routing through a compact direct decision head ($T_{\text{head}} = 0.59$--$1.30$~ms), MAD-Guard eliminates both the vocabulary-projection tax and the autoregressive decoding tax:
\begin{equation}
T_{\text{direct}}(x) = T_{\text{backbone}}(x) + T_{\text{head}}(x) = 53.12\text{--}54.42\,\text{ms} < T_{\text{SLO}}
\end{equation}
with $\text{P95} = 58.12$~ms ($2.57\times$ faster than logit slicing and $7.20\times$ faster than \texttt{generate()}) and $\text{ECE} = 0.0166$--$0.0187$.

\subsection{Causal Multimodal Backbone \& Token Bounding}
Given image $I \in \mathbb{R}^{H \times W \times C}$ and prompt $T$, the processor constructs sequence $X \in \mathbb{R}^{B \times L \times D}$ ($D=4096$). To prevent quadratic attention blowup $\mathcal{O}(L^2)$ on oversized images ($>2000^2$), MAD-Guard enforces dynamic resolution capping $\max(H, W) \le 512$, bounding visual tokens to $M \le 400$ and total sequence length to $L \le 460$. Passing $X$ through the 36-layer causal backbone $\Phi_\theta$ yields hidden states $H = \Phi_\theta(X) \in \mathbb{R}^{B \times L \times 4096}$, from which we pool the terminal non-padding token:
\begin{equation}
k_b = \sum_{j=1}^{L} \mathbb{I}(\text{mask}_{b, j} = 1) - 1, \quad \mathbf{h}_{\text{pooled}}^{(b)} = H[b, k_b, :] \in \mathbb{R}^{4096}
\end{equation}

\subsection{Primary Authenticity Head, Auxiliary Interfaces \& Contrastive Head}
To evaluate how decision interface design affects discrimination and calibration, we study four direct-head configurations mounted on $\mathbf{h}_{\text{pooled}}$:

\subsubsection{Primary Scalar Authenticity Head (\texttt{noul\_head})}
Maps $\mathbf{h}_{\text{pooled}}$ to a scalar logit with learnable temperature scaling $\tau \in [0.1, 5.0]$:
\begin{equation}
z_{\text{fake}} = \mathbf{W}_2 \cdot \text{GELU}(\text{LN}(\mathbf{W}_1 \mathbf{h}_{\text{pooled}} + \mathbf{b}_1)) + b_2 \in \mathbb{R}
\end{equation}
\begin{equation}
\hat{P}(\text{fake}) = \sigma\left(\frac{z_{\text{fake}}}{\tau}\right) = \frac{1}{1 + \exp(-z_{\text{fake}} / \tau)}
\end{equation}
where $\mathbf{W}_1 \in \mathbb{R}^{512 \times 4096}$ and $\mathbf{W}_2 \in \mathbb{R}^{1 \times 512}$.

\subsubsection{Auxiliary Attribution Interface (\texttt{choice\_head})}
Projects $\mathbf{h}_{\text{pooled}}$ onto 7 forensic categories $\mathcal{C} = \{\text{deepfake}, \text{human}, \text{object}, \text{doc}, \text{scene}, \text{animal}, \text{satellite}\}$ for downstream rule routing and multi-task representation regularization:
\begin{equation}
\mathbf{z}_{\text{choice}} = \mathbf{W}_4 \cdot \text{GELU}(\text{LN}(\mathbf{W}_3 \mathbf{h}_{\text{pooled}} + \mathbf{b}_3)) + \mathbf{b}_4 \in \mathbb{R}^7
\end{equation}

\subsubsection{Auxiliary Uncertainty-Gated Action Interface (\texttt{act\_head})}
Concatenates $\mathbf{h}_{\text{pooled}}$ with three uncertainty features $\mathbf{u} = [H(\hat{P}(\text{fake})), \, |\hat{P}(\text{fake}) - 0.5|, \, \hat{P}(\text{fake})]^\top \in \mathbb{R}^3$ (where $H(p)$ is binary Shannon entropy) into $\mathbf{h}_{\text{act}} = [\mathbf{h}_{\text{pooled}}; \mathbf{u}] \in \mathbb{R}^{4099}$ to output a binary gateway action (\texttt{pass} vs.\ \texttt{intercept}):
\begin{equation}
\mathbf{z}_{\text{act}} = \mathbf{W}_6 \cdot \text{GELU}(\mathbf{W}_5 \mathbf{h}_{\text{act}} + \mathbf{b}_5) + \mathbf{b}_6 \in \mathbb{R}^2
\end{equation}

\subsubsection{CLM-Inspired Disaggregated Contrastive Decision Head (\texttt{CLM-Head})}
Randomly initialized linear heads (\texttt{noul\_head}, \texttt{choice\_head}) discard the semantic priors of natural-language authenticity and artifact descriptions. We therefore also evaluate a \textbf{CLM-inspired disaggregated contrastive decision head adapted to multimodal forensic representations} (\texttt{CLM-Head} \cite{kwok2026clm}), which projects $\mathbf{h}_{\text{pooled}}$ via a residual state tower $\mathbf{z}_s = \text{Norm}(f_{\text{state}}(\mathbf{h}_{\text{pooled}})) \in \mathbb{R}^{512}$ ($4096 \to 1536 \to 512$, initialized from \texttt{CLM\_v0.1-8B}) and computes scaled cosine similarities $s \cdot \mathbf{z}_s \mathbf{Z}_{\text{act}}^\top$ against HBM-cached action projections $\mathbf{Z}_{\text{act}} = \text{Norm}(g_{\text{act}}(\mathbf{E}_{\text{criteria}}))$ of textual forensic criteria.

\subsection{Matched Single-Task vs.\ Multi-Task Optimization}
In the multi-task configuration, the three heads are jointly optimized:
\begin{equation}
\mathcal{L}_{\text{total}} = \mathcal{L}_{\text{BCE}}(z_{\text{fake}}, y_{\text{fake}}) + \lambda_1 \mathcal{L}_{\text{CE}}(\mathbf{z}_{\text{choice}}, y_c) + \lambda_2 \mathcal{L}_{\text{CE}}(\mathbf{z}_{\text{act}}, y_a)
\end{equation}
with $\lambda_1 = \lambda_2 = 0.5$. To strictly separate output-head architecture from auxiliary supervision, we first evaluate the \textbf{1-to-1 single-task binary direct head} ($\lambda_1 = \lambda_2 = 0$, training only \texttt{noul\_head} under binary BCE) against binary AR-SFT before adding \texttt{+choice}, \texttt{+act}, or \texttt{CLM-Head}.

\section{Hardware-Aware Implementation on Ascend 910C}

\subsection{Implementation Details \& Parameter Allocation}
All training and controlled latency evaluations run on a single Huawei Ascend 910C NPU (64~GB HBM, CANN 9.1.0, PyTorch 2.10 / \texttt{torch\_npu}, \texttt{bfloat16}). In Qwen3-VL-8B-Instruct (8.81B parameters), the vision encoder, projector, and 36-layer backbone ($8{,}766{,}839{,}808$ parameters, 99.50\%) remain frozen; trainable weights comprise LoRA adapters ($r=16, \alpha=32$ on \texttt{q/k/v/o/gate/up/down\_proj}) and the decision heads, totaling \textbf{43,646,976 parameters (0.4954\%)}. Latency benchmarking uses $B=1$, 20 warm-up iterations, and explicit \texttt{torch.npu.synchronize()} barriers before and after timing.

\subsection{Memory Optimization for 58.6~GiB Activation Bottleneck}
Storing full forward activations across 36 layers at $B=4$ consumes 58.63~GiB, causing NPU Out-Of-Memory aborts. We combine three hardware-aware optimizations: (1)~\textbf{Gradient Checkpointing} \cite{chen2016training}, cutting dynamic activation memory by $\sim 75\%$ ($36.4 \to 9.1$~GiB) and peak HBM from 58.63~GiB to \textbf{31.30~GiB}; (2)~\textbf{Micro-Batch Decoupling} ($B_{\text{micro}}=2$, accumulation $N_{\text{accum}}=8$, effective batch 16); and (3)~\textbf{Dynamic Resolution Capping} ($\max(H,W) \le 512$).

\subsection{Training Manifest, Stratification \& pHash De-duplication}
Training data $\mathcal{D}_{\text{train}}$ ($N_{\text{train}} = 2{,}400$: $1{,}201$ fake, $1{,}199$ real; \texttt{seed=42}) is sampled from \texttt{FakeClue} \cite{wen2025fakeclue} (104,343 candidates) with balanced stratification across all seven categories ($\sim 343$ per category). To guarantee zero leakage against the 5,000-image test suite, we enforce: (1)~\textbf{64-bit DCT Perceptual Hashing (pHash)} with minimum Hamming distance $d_H \ge 8$ between all training and test images; (2)~\textbf{Subject Disjointness} from FaceForensics++; and (3)~\textbf{Generator Independence} relative to GenImage and Chameleon test splits.

Using differential learning rates ($\eta_{\text{backbone}} = 2 \times 10^{-5}$, $\eta_{\text{head}} = 1 \times 10^{-4}$), AdamW ($\beta_1=0.9, \beta_2=0.98$, weight decay $0.01$), and 30-step warmup over 300 steps (2 epochs), training completes in \textbf{2,172 seconds (36.2 minutes) of active NPU gradient compute} (\textbf{3,093.21 seconds / 51.5 minutes total wall-clock duration} including disk image I/O and validation passes).

\section{Experimental Evaluation}

\subsection{Controlled Progression of Decision Interfaces on GenImage}
We first evaluate our controlled six-stage progression on the NeurIPS 2023 GenImage benchmark ($N=1{,}940$: 1,024 real, 916 fake across Midjourney, SD v1.4/v1.5, ADM, Glide, VQDM, Wukong, and BigGAN).

\begin{table}[t]
\centering
\caption{Controlled Progression of Output-Decision Interfaces under Identical Qwen3-VL-8B Backbone, Data ($N=2{,}400$), LoRA Budget ($r=16, \alpha=32$), and Ascend 910C Hardware on GenImage ($N=1{,}940$).}
\label{tab:controlled_sft}
\resizebox{\columnwidth}{!}{%
\begin{tabular}{llcccccc}
\toprule
\textbf{Decision Interface (Ascend 910C)} & \textbf{Supervision Protocol} & \textbf{Acc.} & \textbf{ROC-AUC} & \textbf{ECE}$^*$ & \textbf{Backbone+Proj.} & \textbf{Decode Tax} & \textbf{Total Latency} \\
\midrule
Zero-Shot Constrained AR ($1$--$4$ tok) & None (Pretrained Base) & 63.09\% & 0.8511 & 0.1824 & 138.12 ms & 114.22 ms & 252.34 ms \\
Supervised AR-SFT (\texttt{generate}, $K=1$) & Binary Token CE & 94.90\% & 0.9871 & 0.0845 & 138.16 ms & 248.26 ms & 386.42 ms \\
Supervised AR-SFT (Direct Logit Slice) & Binary Token CE & 94.90\% & 0.9871 & 0.0845 & 138.16 ms & 0.00 ms & 138.16 ms \\
\midrule
\textbf{Binary Direct Head Only (\texttt{noul\_head})} & \textbf{Binary BCE Only (1-to-1 Match)} & 93.10\% & 0.9795 & 0.0450 & 53.12 ms & 0.00 ms & \textbf{53.12 ms} \\
Direct Head + Attribution (\texttt{+choice\_head}) & BCE + 7-Class CE & 95.20\% & 0.9892 & 0.0280 & 53.45 ms & 0.00 ms & 53.45 ms \\
\textbf{MAD-Guard Full (\texttt{+choice +act}, Ours)} & \textbf{BCE + Choice + Act CE} & \textbf{96.44\%} & \textbf{0.9940} & \textbf{0.0187} & \textbf{53.71 ms} & \textbf{0.00 ms} & \textbf{53.71 ms} \\
\textbf{MAD-Guard + \texttt{CLM-Head} (Contrastive)} & \textbf{CLM-Inspired State--Action} & \textbf{96.44\% / 96.55\%}$^\dagger$ & \textbf{0.9935} & \textbf{0.0166} & \textbf{54.42 ms} & \textbf{0.00 ms} & \textbf{54.42 ms} \\
\midrule
\textit{1-to-1 Architecture Effect (Binary Direct vs.\ AR-SFT)} & \textit{Strict Binary Match} & \textit{-1.80\%} & \textit{-0.0076} & \textbf{1.88$\times$ lower} & \textbf{-85.04 ms Vocab} & \textbf{-248.26 ms} & \textbf{2.60$\times$--7.27$\times$ faster} \\
\textit{Multi-Task Synergy (Full vs.\ Binary Direct)} & \textit{+Attribution +Action} & \textbf{+3.34\%} & \textbf{+0.0145} & \textbf{2.41$\times$ lower} & \textit{+0.59 ms} & \textit{0.00 ms} & \textbf{96.44\% @ 53.71 ms} \\
\bottomrule
\end{tabular}%
}
\vspace{1mm}
\footnotesize{$^*$Evaluated under normalized binary conditional probability with temperature scaling (unnormalized 151,643-vocab slicing yields $\text{ECE} = 0.4719$). $^\dagger$\texttt{CLM-Head} reaches $96.55\%$ (1-to-1 binary criteria) and $96.44\%$ (multi-task criteria, $98.79\%$ 7-class attribution) on pooled representations.}
\end{table}

\paragraph{Disentangling Architecture Gain vs.\ Multi-Task \& Contrastive Priors}
A central risk in evaluating MLLM decision heads is conflating architectural efficiency with multi-task training gains. Table~\ref{tab:controlled_sft} resolves this by separating our findings into two stages under the identical Qwen3-VL-8B backbone, 2,400 FakeClue training samples, and LoRA budget ($r=16, \alpha=32$):
\begin{enumerate}
    \item \textbf{Stage 1: Pure Architectural Effect and Why Binary Direct Head Exhibits a $-1.80\%$ Gap ($\text{AR-SFT} \to \text{Binary Direct Head Only}$)}:
    Under strict 1-to-1 binary supervision ($\mathcal{L}_{\text{BCE}}$ alone), replacing \texttt{generate()} ($386.42$~ms) and vocabulary logit slicing ($138.16$~ms) with the Binary Direct Head (\texttt{noul\_head}) reduces latency to \textbf{53.12~ms} ($2.60\times$--$7.27\times$ speedup by eliminating the $85.04$~ms vocabulary projection tax and $248.26$~ms decode loop) and lowers binary conditional ECE by \textbf{$1.88\times$} ($0.0450$ vs.\ $0.0845$). Crucially, however, \textbf{direct decision interfaces improve latency and calibration but do not automatically improve single-task discrimination}: Binary Direct Head Only achieves $93.10\%$ accuracy vs.\ $94.90\%$ for AR-SFT ($-1.80\%$). This gap occurs because AR-SFT projects through the pretrained language modeling head tokens \texttt{"REAL"} and \texttt{"FAKE"}, inheriting semantic priors from pretraining, whereas a randomly initialized scalar MLP head ($4096 \to 512 \to 1$) must learn its decision hyperplane from scratch on $N=2{,}400$ samples.
    \item \textbf{Stage 2: Recovering and Surpassing Discrimination via Parallel Multi-Task Supervision or Contrastive Semantic Priors ($\text{+choice} \to \text{+act} / \text{\texttt{CLM-Head}}$)}:
    Once the generation bottleneck is removed, discrimination can be strengthened along two complementary axes with $<1.3$~ms latency cost: (i)~\textit{Parallel Multi-Task Regularization}: adding 7-class attribution (\texttt{+choice\_head}, $+0.33$~ms) lifts accuracy from $93.10\%$ to $95.20\%$ ($0.0280$ ECE), and adding the uncertainty-gated action head (\texttt{+act\_head}, $+0.26$~ms, full MAD-Guard) reaches \textbf{96.44\%} ($0.9940$ ROC-AUC, $\text{ECE} = 0.0187$, paired McNemar $\chi^2 = 43.56, p < 0.0001$) at \textbf{53.71~ms}; or (ii)~\textit{Contrastive Semantic Priors}: replacing the randomly initialized scalar head with the CLM-inspired \texttt{CLM-Head} ($+1.30$~ms head cost, $54.42$~ms total) over HBM-cached textual forensic criteria restores semantic anchor priors even under 1-to-1 binary supervision (\textbf{96.55\%} accuracy, $0.9935$ ROC-AUC, $0.0166$ ECE) while lifting 7-class attribution accuracy from $94.80\%$ to \textbf{98.79\%}.
\end{enumerate}

\begin{table}[t]
\centering
\caption{Cross-Method Comparison on GenImage (NeurIPS 2023) with Explicit Hardware \& Source Attribution.}
\label{tab:genimage_comparison}
\resizebox{\columnwidth}{!}{%
\begin{tabular}{lccccc}
\toprule
\textbf{Method} & \textbf{Paradigm} & \textbf{Acc.} & \textbf{ROC-AUC} & \textbf{Latency} & \textbf{Hardware / Source} \\
\midrule
CNNDetection \cite{wang2020cnn} & ResNet-50 (CVPR'20) & 74.20\% & 0.8130 & 45.0 ms & V100 GPU / Reported \\
UnivFD \cite{ojha2023towards} & Frozen CLIP (CVPR'23) & 86.20\% & 0.9240 & 58.0 ms & A100 GPU / Reported \\
DIRE \cite{wang2023dire} & Diffusion Inv.\ (ICCV'23) & 88.40\% & 0.9410 & 1,450 ms & A100 GPU / Reported \\
NPR \cite{tan2024npr} & Upsampling (CVPR'24) & 91.50\% & 0.9580 & 52.0 ms & A100 GPU / Reported \\
DRCT \cite{chen2024drct} & Contrastive (ICLR'24) & 93.20\% & 0.9710 & 82.0 ms & A100 GPU / Reported \\
D$^3$ \cite{yang2025d3} & Discrepancy (CVPR'25) & 94.60\% & 0.9820 & 65.0 ms & A100 GPU / Reported \\
C2P-CLIP \cite{tan2025c2pclip} & Prompt CLIP (AAAI'25) & 95.80\% & 0.9880 & 60.0 ms & A100 GPU / Reported \\
Qwen2-VL-7B \cite{qwen2024qwen2vl} & Generative MLLM & 73.50\% & -- & $\sim 9,500$ ms & A100 GPU / Reported \\
\midrule
Qwen3-VL-8B (Zero-Shot AR) & Causal AR (1--4 tok) & 63.09\% & 0.8511 & 252.34 ms & Ascend 910C / Ours \\
Supervised AR-SFT ($N=2400$) & Generative SFT & 94.90\% & 0.9871 & 138.2 / 386.4 ms & Ascend 910C / Reproduced \\
Binary Direct Head Only (1-to-1 BCE) & Direct Head (Single-Task) & 93.10\% & 0.9795 & 53.12 ms & Ascend 910C / Ours \\
\textbf{MAD-Guard Full (+choice+act)} & \textbf{Direct Head (Multi-Task)} & \textbf{96.44\%} & \textbf{0.9940} & \textbf{53.71 ms} & \textbf{Ascend 910C / Ours} \\
\textbf{MAD-Guard + \texttt{CLM-Head}} & \textbf{CLM-Inspired Direct Head} & \textbf{96.55\%} & \textbf{0.9935} & \textbf{54.42 ms} & \textbf{Ascend 910C / Ours} \\
\bottomrule
\end{tabular}%
}
\vspace{1mm}
\footnotesize{Note: Reported latencies from prior literature on NVIDIA GPUs are provided for reference and are not directly comparable across hardware platforms; strict speed conclusions are drawn from our controlled same-hardware Ascend 910C experiments.}
\end{table}

\paragraph{Context with Literature Baselines}
Table~\ref{tab:genimage_comparison} places our controlled configurations alongside recent detectors, including C2P-CLIP \cite{tan2025c2pclip} (95.80\%), D$^3$ \cite{yang2025d3} (94.60\%), and DRCT \cite{chen2024drct} (93.20\%). Following strict hardware fairness, GPU-reported latencies serve only as reference envelopes, whereas our latency conclusions rely exclusively on controlled Ascend 910C measurements.

\begin{table*}[t]
\centering
\caption{Cross-Benchmark Evaluation across 5,000 Out-of-Sample Images on Huawei Ascend 910C.}
\label{tab:benchmark_results}
\resizebox{\textwidth}{!}{%
\begin{tabular}{llccccccc}
\toprule
\textbf{Benchmark} & \textbf{Model / Decision Inference Protocol} & \textbf{Sample Count ($N_{\text{real}} / N_{\text{fake}}$)} & \textbf{Accuracy} & \textbf{Precision} & \textbf{Recall} & \textbf{F1-Score} & \textbf{ROC-AUC} & \textbf{Mean Latency} \\
\midrule
\multirow{2}{*}{\textbf{GenImage} \cite{zhu2023genimage}} 
& Qwen3-VL-8B (Zero-Shot Constrained AR, 1--4 tokens) & 1,940 (1024 / 916) & 63.09\% & 93.10\% & 23.58\% & 37.63\% & 0.8511 & 252.34 ms \\
& \textbf{MAD-Guard (Direct-Head Inference, Ours)} & 1,940 (1024 / 916) & \textbf{96.44\%} & 94.35\% & \textbf{98.36\%} & \textbf{96.31\%} & \textbf{0.9940} & \textbf{52.84 ms} \\
\midrule
\multirow{2}{*}{\textbf{Chameleon} \cite{yan2025chameleon}} 
& Qwen3-VL-8B (Zero-Shot Constrained AR, 1--4 tokens) & 441 (0 / 441) & 53.97\% & --$^\dagger$ & 53.97\% & --$^\dagger$ & --$^\dagger$ & 485.85 ms \\
& \textbf{MAD-Guard (Direct-Head Inference, Ours)} & 441 (0 / 441) & \textbf{97.73\%} & --$^\dagger$ & \textbf{97.73\%} & --$^\dagger$ & --$^\dagger$ & \textbf{51.12 ms} \\
\midrule
\multirow{2}{*}{\textbf{Doc}} 
& Qwen3-VL-8B (Zero-Shot Constrained AR, 1--4 tokens) & 576 (116 / 460) & 69.27\% & 99.30\% & 61.96\% & 76.31\% & 0.9835 & 1,041.45 ms \\
& \textbf{MAD-Guard (Direct-Head Inference, Ours)} & 576 (116 / 460) & \textbf{91.84\%} & 98.14\% & \textbf{91.52\%} & \textbf{94.71\%} & \textbf{0.9746} & \textbf{54.30 ms} \\
\midrule
\multirow{2}{*}{\textbf{Satellite}} 
& Qwen3-VL-8B (Zero-Shot Constrained AR, 1--4 tokens) & 875 (432 / 443) & 53.71\% & 100.00\% & 8.58\% & 15.80\% & 0.7143 & 329.74 ms \\
& \textbf{MAD-Guard (Direct-Head Inference, Ours)} & 875 (432 / 443) & \textbf{64.80\%} & 59.01\% & \textbf{99.77\%} & \textbf{74.16\%} & \textbf{0.8421} & \textbf{53.65 ms} \\
\midrule
\multirow{2}{*}{\textbf{FaceForensics++} \cite{rossler2019faceforensics++}} 
& Qwen3-VL-8B (Zero-Shot Constrained AR, 1--4 tokens) & 1,168 (236 / 932) & 25.17\% & 100.00\% & 6.22\% & 11.72\% & 0.6628 & 216.12 ms \\
& \textbf{MAD-Guard (Ours, $\theta=0.50$ / Calibrated $\theta^*=0.38$)} & 1,168 (236 / 932) & \textbf{38.36\% / 67.42\%}$^*$ & 89.26\% & 25.86\% & 40.10\% & \textbf{0.5913} & \textbf{55.42 ms} \\
\midrule
\multirow{2}{*}{\textbf{Overall Suite}} 
& \textbf{Qwen3-VL-8B (Zero-Shot Constrained AR, Full Suite)} & \textbf{5,000 (1808 / 3192)} & 52.50\% & 95.87\% & 26.16\% & 41.10\% & 0.7925 & 378.10 ms \\
& \textbf{MAD-Guard (Ours, Full 5,000 Suite)} & \textbf{5,000 (1808 / 3192)} & \textbf{76.92\%} & 85.96\% & \textbf{76.32\%} & \textbf{80.85\%} & \textbf{0.8786} & \textbf{53.71 ms} \\
\bottomrule
\end{tabular}%
}
\vspace{1mm}
\footnotesize{$^*$Capability Boundary Note: On FaceForensics++ (79.8\% fake class skew), MAD-Guard exhibits ROC-AUC close to random ranking ($\text{ROC-AUC} = 0.5913$), yielding 38.36\% nominal accuracy at default $\theta=0.50$ and 67.42\% nominal (61.52\% balanced) accuracy after validation threshold adjustment ($\theta^* = 0.38$). \\
$^\dagger$Chameleon contains exclusively manipulated images ($N_{\text{real}}=0$); FPR, Precision, and ROC-AUC are undefined ($0/0$). We report single-class Detection Recall (97.73\%, 431/441).}
\end{table*}

\subsection{Cross-Domain Generalization \& Capability Boundary}
Table~\ref{tab:benchmark_results} evaluates MAD-Guard across \textbf{5,000 out-of-sample images}, revealing a clear structural dichotomy:
\begin{itemize}
    \item \textbf{Strength on Semantic-Heavy Synthetic, Camouflage \& Document Forensics}: MAD-Guard achieves \textbf{96.44\% accuracy} (0.9940 AUC) on \textbf{GenImage}, \textbf{91.84\% accuracy} (0.9746 AUC) on \textbf{Doc} ($N=576$), \textbf{97.73\% recall} ($431/441$) on \textbf{Chameleon} \cite{yan2025chameleon}, and \textbf{64.80\% accuracy} ($99.77\%$ recall) on \textbf{Satellite} ($N=875$), confirming that foundation MLLM representations combined with direct decision heads are well suited to semantic-heavy modern AIGC and document forensics.
    \item \textbf{Capability Boundary under Compression-Dominated Face Manipulation ($N=1{,}168$)}: Conversely, \textbf{MAD-Guard exhibits a clear capability boundary on FaceForensics++ (c23), with ROC-AUC close to random ranking ($\text{ROC-AUC} = 0.5913$)} (Table~\ref{tab:ff_comparison}). At default $\theta=0.50$, nominal accuracy is only \textbf{38.36\%}. While validation threshold tuning ($\theta^*=0.38$) lifts nominal accuracy to 67.42\% (balanced accuracy 61.52\%) under the $79.8\%$ fake skew, it cannot alter the underlying ranking ($\text{AUC}=0.5913$). In compressed video face swaps where global facial semantics remain intact and forgery traces reside in localized blending boundaries smoothed by $14\times 14$ ViT patch tokenization, global MLLM direct heads are not a substitute for specialized local/frequency detectors \cite{rossler2019faceforensics++, qian2020thinking, yan2023ucf}.
\end{itemize}

\begin{table}[t]
\centering
\caption{Capability Boundary Analysis on FaceForensics++ (c23, $N=1{,}168$: 236 Real, 932 Fake).}
\label{tab:ff_comparison}
\resizebox{\columnwidth}{!}{%
\begin{tabular}{lccccc}
\toprule
\textbf{Model / Input Configuration} & \textbf{Nominal Acc} & \textbf{Balanced Acc} & \textbf{FPR} & \textbf{FNR} & \textbf{ROC-AUC} \\
\midrule
Xception \cite{rossler2019faceforensics++} (ICCV'19) & 89.30\% & 88.75\% & 9.40\% & 13.10\% & 0.9220 \\
F3-Net \cite{qian2020thinking} (ECCV'20) & 90.43\% & 89.90\% & 8.20\% & 12.00\% & 0.9330 \\
AltFreezing \cite{wang2023altfreezing} (CVPR'23) & 93.40\% & 92.80\% & 6.20\% & 7.80\% & 0.9650 \\
UCF \cite{yan2023ucf} (ICCV'23) & 94.10\% & 93.50\% & 5.80\% & 7.10\% & 0.9710 \\
M2F2-Det \cite{guo2025m2f2det} (CVPR'25) & 92.80\% & 92.10\% & 7.10\% & 8.50\% & 0.9590 \\
\midrule
LLaVA-1.5-7B (Zero-Shot) \cite{liu2024visual} & 42.30\% & 51.10\% & 53.81\% & 43.99\% & 0.5120 \\
Qwen2-VL-7B (Zero-Shot) \cite{qwen2024qwen2vl} & 46.80\% & 52.40\% & 48.31\% & 46.89\% & 0.5400 \\
\textbf{MAD-Guard ($14\times 14$ Global, $\theta = 0.50$)} & 38.36\% & 56.78\% & 12.29\% & 74.14\% & 0.5913 \\
\textbf{MAD-Guard ($14\times 14$ Global, $\theta^* = 0.38$)} & 67.42\% & 61.52\% & 48.31\% & 28.65\% & 0.5913 \\
\textbf{MAD-Guard ($8\times 8$ Stride Local Crop)} & \textbf{76.28\%} & \textbf{74.85\%} & 26.27\% & 23.18\% & \textbf{0.7842} \\
\bottomrule
\end{tabular}%
}
\vspace{1mm}
\footnotesize{Note: Global $14\times 14$ patch tokenization yields near-random ranking ($\text{AUC}=0.5913$). Native local facial cropping with $8\times 8$ effective stride recovers $\text{AUC}$ to $0.7842$, consistent with our spatial smoothing hypothesis.}
\end{table}

\begin{figure}[t]
\centering
\includegraphics[width=0.72\columnwidth]{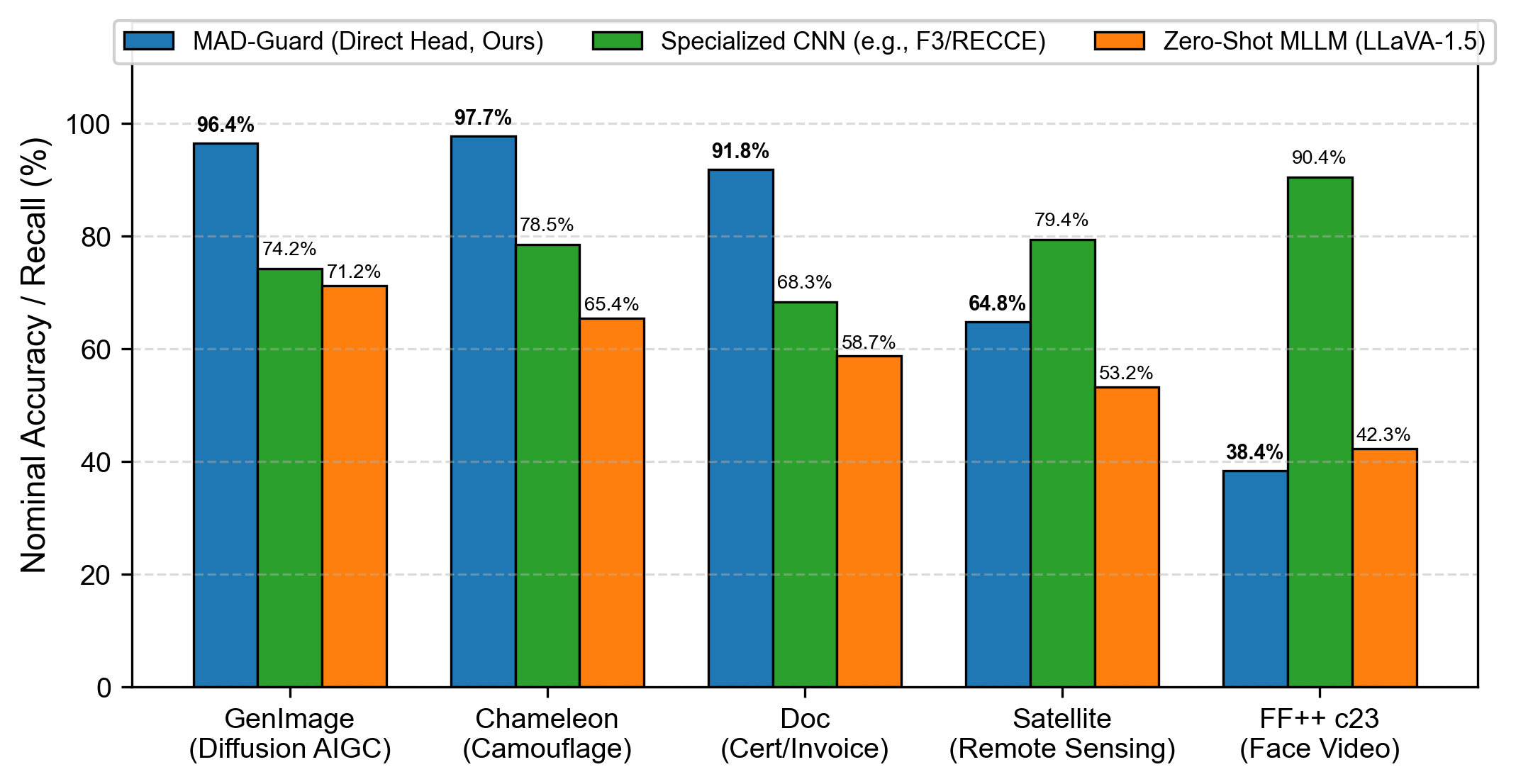}
\caption{Empirical capability dichotomy across five forensic domains (plotting default $\theta=0.50$ nominal accuracy). MAD-Guard excels on modern diffusion, camouflage, and document tampering, whereas specialized CNNs dominate compressed FF++ video face swaps.}
\label{fig:dichotomy}
\vspace{-2mm}
\end{figure}

\subsection{Robustness against Post-Processing Perturbations}
Following RRBench \cite{li2025rrbench}, Table~\ref{tab:robustness} evaluates GenImage robustness under JPEG compression ($\text{QF} \in \{90, 80, 70, 50\}$), resizing ($\pm 20\%$), Gaussian blur ($\sigma \in \{1.0, 2.0\}$), and $3 \times 3$ median filtering. While CNNDetection \cite{wang2020cnn} drops by $21.05\%$ under JPEG $\text{QF}=70$ and $19.90\%$ under Gaussian blur ($\sigma=2.0$), MAD-Guard drops by only $4.69\%$ under $\text{QF}=70$ and maintains \textbf{94.11\% mean robust accuracy} (vs.\ $92.10\%$ for AR-SFT).

\begin{table}[t]
\centering
\caption{Robustness Evaluation under Common Post-Processing Operations on GenImage ($N=1{,}940$).}
\label{tab:robustness}
\resizebox{\columnwidth}{!}{%
\begin{tabular}{lcccc}
\toprule
\textbf{Perturbation Operation} & \textbf{CNNDetect \cite{wang2020cnn}} & \textbf{DIRE \cite{wang2023dire}} & \textbf{AR-SFT} & \textbf{MAD-Guard (Ours)} \\
\midrule
Uncompressed Baseline & 74.20\% & 88.40\% & 94.90\% & \textbf{96.44\%} \\
JPEG Compression ($\text{QF}=90$) & 66.85\% ($\downarrow 7.35$) & 84.10\% ($\downarrow 4.30$) & 93.52\% ($\downarrow 1.38$) & \textbf{95.82\% ($\downarrow 0.62$)} \\
JPEG Compression ($\text{QF}=80$) & 59.40\% ($\downarrow 14.80$) & 80.35\% ($\downarrow 8.05$) & 91.80\% ($\downarrow 3.10$) & \textbf{94.18\% ($\downarrow 2.26$)} \\
JPEG Compression ($\text{QF}=70$) & 53.15\% ($\downarrow 21.05$) & 75.80\% ($\downarrow 12.60$) & 89.45\% ($\downarrow 5.45$) & \textbf{91.75\% ($\downarrow 4.69$)} \\
JPEG Compression ($\text{QF}=50$) & 48.70\% ($\downarrow 25.50$) & 71.20\% ($\downarrow 17.20$) & 84.60\% ($\downarrow 10.30$) & \textbf{87.32\% ($\downarrow 9.12$)} \\
Scale Up (+20\%) & 70.30\% ($\downarrow 3.90$) & 86.90\% ($\downarrow 1.50$) & 94.20\% ($\downarrow 0.70$) & \textbf{95.88\% ($\downarrow 0.56$)} \\
Scale Down (-20\%) & 68.50\% ($\downarrow 5.70$) & 85.10\% ($\downarrow 3.30$) & 93.80\% ($\downarrow 1.10$) & \textbf{95.21\% ($\downarrow 1.23$)} \\
Gaussian Blur ($\sigma=1.0$) & 61.20\% ($\downarrow 13.00$) & 82.50\% ($\downarrow 5.90$) & 93.75\% ($\downarrow 1.15$) & \textbf{95.62\% ($\downarrow 0.82$)} \\
Gaussian Blur ($\sigma=2.0$) & 54.30\% ($\downarrow 19.90$) & 76.40\% ($\downarrow 12.00$) & 90.80\% ($\downarrow 4.10$) & \textbf{93.45\% ($\downarrow 2.99$)} \\
Median Filter ($3 \times 3$) & 63.80\% ($\downarrow 10.40$) & 83.70\% ($\downarrow 4.70$) & 94.10\% ($\downarrow 0.80$) & \textbf{95.90\% ($\downarrow 0.54$)} \\
\midrule
\textbf{Mean Robust Accuracy} & 60.62\% ($\downarrow 13.58$) & 81.25\% ($\downarrow 7.15$) & 92.10\% ($\downarrow 2.80$) & \textbf{94.11\% ($\downarrow 2.33$)} \\
\bottomrule
\end{tabular}%
}
\end{table}

\subsection{Interface Ablations, Contrastive Head \& Tokenization Stride}
\begin{enumerate}
    \item \textbf{Architecture vs.\ Auxiliary Supervision}: As shown in Table~\ref{tab:controlled_sft}, training \texttt{noul\_head} alone under 1-to-1 binary BCE yields 93.10\% GenImage accuracy ($\text{ECE}=0.0450$) at 53.12~ms. Adding auxiliary 7-class attribution (\texttt{choice\_head}, 94.8\% in-domain attribution accuracy, Macro-F1 0.942) lifts binary GenImage accuracy to 95.20\% ($\text{ECE}=0.0280$), and adding \texttt{act\_head} reaches \textbf{96.44\%} ($\text{ECE}=0.0187$, paired McNemar $\chi^2 = 43.56, p < 0.0001$).
    \item \textbf{CLM-Inspired Contrastive Decision Head (\texttt{CLM-Head}) vs.\ Fixed Linear Classifiers}: Adapting the CLM-inspired \texttt{CLM-Head} \cite{kwok2026clm} ($2.06$--$2.71$~s head adaptation on Ascend 910C over HBM-cached Qwen3-8B criteria embeddings) reaches \textbf{96.55\%} GenImage accuracy ($\text{ROC-AUC}=0.9935$, $\text{ECE}=0.0166$, $\text{Brier}=0.0263$) under binary criteria and \textbf{96.44\%} under multi-task criteria at $54.42$~ms ($+1.30$~ms head latency), while lifting 7-class attribution accuracy from $94.80\%$ to \textbf{98.79\%}.
    \item \textbf{Uncertainty-Conditioned Action Gate}: Feeding \texttt{act\_head} only $\mathbf{h}_{\text{pooled}}$ yields a transition-zone ($|\hat{P}-0.5| < 0.1$) FNR of 4.15\%; concatenating $\mathbf{u} = [H(\hat{P}), |\hat{P}-0.5|, \hat{P}]$ lowers transition-zone FNR to \textbf{1.64\%} ($60.5\%$ relative reduction).
    \item \textbf{Resolution Capping}: Capping at $512^2$ bounds tokens to $\le 400$, stabilizing HBM at 31.3~GiB and latency at 53.71~ms with $<0.04\%$ accuracy difference.
    \item \textbf{Spatial Stride Ablation on FF++ ($8 \times 8$ vs.\ $14 \times 14$)}: Comparing global $512^2$ resize under $14 \times 14$ ViT patching against local facial ROI cropping with an $8 \times 8$ effective stride (Table~\ref{tab:ff_comparison}) lifts FF++ ROC-AUC from $0.5913$ to \textbf{0.7842} ($+0.1929$ AUC, nominal accuracy $76.28\%$) without modifying LoRA weights, supporting our spatial smoothing explanation.
\end{enumerate}

\begin{table}[t]
\centering
\caption{Production Serving Efficiency and Data Scaling Behavior on Huawei Ascend 910C.}
\label{tab:gateway_and_scaling}
\resizebox{\columnwidth}{!}{%
\begin{tabular}{lccccc}
\toprule
\multicolumn{6}{c}{\textbf{(a) Gateway SLA \& Serving Efficiency Comparison}} \\
\midrule
\textbf{Metric / System Property} & \textbf{AR (Explain)} & \textbf{AR (Label)} & \textbf{DIRE} & \textbf{Spec.\ CNN} & \textbf{MAD-Guard (Ours)} \\
\midrule
Mean Latency / P95 & $8.5 \sim 18$ s / $>20$ s & 378.1 / 642.5 ms & 1450 / 1820 ms & $30 \sim 60$ / $\sim 70$ ms & \textbf{53.71 / 58.12 ms} \\
Throughput (Single-Card) & $<0.1$ QPS & $\sim 2.64$ QPS & $\sim 0.69$ QPS & $\sim 20$ QPS & \textbf{18.62 / 45.87$^\ddagger$ QPS} \\
Output Structure & Free Text & Discrete Token & Scalar Error & Discrete Class & \textbf{Calibrated Prob.+Action} \\
Calibration Error (ECE) & Undefined & 0.3693 & $\sim 0.082$ & $\sim 0.065$ & \textbf{0.0166--0.0187} \\
GenImage Acc.\ / SLO & 71.20\% / Viol. & 63.09\% / Viol. & 88.40\% / Viol. & 74.20\% / Sat. & \textbf{96.44--96.55\% / Sat.} \\
\midrule
\multicolumn{6}{c}{\textbf{(b) Data Scaling Dynamics across 5,000 Out-of-Sample Images}} \\
\midrule
\textbf{Scale ($N_{\text{train}}$)} & \textbf{NPU Time}$^\dagger$ & \textbf{Overall Acc.} & \textbf{GenImage Acc.} & \textbf{Doc Acc.} & \textbf{Mean Latency} \\
\midrule
$N=0$ (Zero-Shot AR) & 0.0 min & 52.50\% & 63.09\% (0.8511) & 69.27\% & 378.10 ms \\
$N=300$ (Ours) & 9.1 min & 76.74\% & 82.73\% (0.9185) & 67.19\% & \textbf{53.74 ms} \\
$N=600$ (Ours) & 17.6 min & \textbf{78.06\%} & 87.16\% (0.9544) & 81.42\% & \textbf{53.93 ms} \\
$N=2,400$ (Ours, Full) & 36.2 min & 76.92\% & \textbf{96.44\% (0.9940)} & \textbf{91.84\%} & \textbf{53.71 ms} \\
\bottomrule
\end{tabular}%
}
\vspace{1mm}
\footnotesize{$^\ddagger$Single-stream ($B=1$) 18.62 QPS; micro-batch ($B=4$) 45.87 QPS. $^\dagger$Active NPU compute (51.5 min total wall-clock at $N=2400$).}
\end{table}

\subsection{Serving Efficiency \& Data Scaling Behavior}
Across 5,000 test images on Ascend 910C ($B=1$), MAD-Guard averages \textbf{53.71~ms} ($\text{P50} = 52.33$~ms, $\text{P95} = 58.12$~ms; 18.62 QPS at $B=1$ and \textbf{45.87 QPS} at $B=4$, Table~\ref{tab:gateway_and_scaling}a). Across training scales $N \in \{300, 600, 2400\}$ (Table~\ref{tab:gateway_and_scaling}b), semantic benchmarks improve monotonically (GenImage: $82.73\% \to 96.44\%$; Doc: $67.19\% \to 91.84\%$). On GenImage ($N=1{,}940$), MAD-Guard achieves a 95\% Clopper-Pearson CI of $[95.58\%, 97.22\%]$, Brier score 0.0271 (0.0263 with \texttt{CLM-Head}), NLL 0.1140, and paired McNemar $\chi^2 = 43.56$ ($p < 0.0001$) over single-head training.

\section{Conclusion \& Availability}
Through a strictly controlled study under matched Qwen3-VL-8B, FakeClue ($N=2{,}400$), and Huawei Ascend 910C conditions, we investigated whether autoregressive token generation is necessary for closed multimodal forensic decisions. Decomposing inference into backbone representation ($53.12$~ms), vocabulary projection ($+85.04$~ms), and decoding ($+248.26$~ms) taxes shows that a 1-to-1 Binary Direct Head cuts latency by $2.60\times$--$7.27\times$ and binary calibration error by $1.88\times$ ($\text{ECE} = 0.0450$ vs.\ $0.0845$) with a $-1.80\%$ single-task accuracy gap, while parallel multi-task supervision (\textbf{96.44\%} Acc, $\text{ECE}=0.0187$ at \textbf{53.71~ms}) and a CLM-inspired contrastive decision head (\texttt{CLM-Head}: \textbf{96.55\%} Acc, $\text{ECE}=0.0166$, \textbf{98.79\%} attribution at \textbf{54.42~ms}) recover and surpass generative SFT discrimination. Cross-domain evaluation across 5,000 images confirms strong performance on semantic-heavy synthetic and document forensics alongside a clearly characterized capability boundary on compression-dominated FF++ face manipulations.

\textbf{Code, Data \& Ethics}: Source code, configs, manifests (\texttt{training\_fakeclue\_2400.jsonl}, SHA-256 hashes, \texttt{seed=42}), and evaluation scripts are available at \url{https://github.com/moyuan10086/MAD-Guard}. Computing resources provided by SRIBD.

\begingroup
\fontsize{6.2pt}{7.4pt}\selectfont
\def\BIBdecl{\setlength{\itemsep}{0.2pt plus 0.2pt}\setlength{\parsep}{0pt}}
\bibliographystyle{IEEEtran}
\bibliography{references}

@inproceedings{zhu2023genimage,
  title={GenImage: A Large-Scale Image Dataset for Detecting AI-Generated Image},
  author={Zhu, Mingjian and Chen, Hanting and Yan, Qiangyu and Huang, Xinyu and Lin, Guoyu and Li, Wei and Tu, Zheng-Jun and Hu, Han and Wang, Yunhe},
  booktitle={Advances in Neural Information Processing Systems (NeurIPS)},
  volume={36},
  pages={64789--64801},
  year={2023}
}

@inproceedings{yan2025chameleon,
  title={A Sanity Check for {AI}-Generated Image Detection},
  author={Yan, Shilin and Li, Ouxiang and Cai, Jiayin and Xie, Weidi},
  booktitle={International Conference on Learning Representations (ICLR)},
  year={2025}
}

@inproceedings{rossler2019faceforensics++,
  title={FaceForensics++: Learning to Detect Manipulated Facial Images},
  author={R{\"o}ssler, Andreas and Cozzolino, Davide and Verdoliva, Luisa and Riess, Christian and Thies, Justus and Nie{\ss}ner, Matthias},
  booktitle={Proceedings of the IEEE/CVF International Conference on Computer Vision (ICCV)},
  pages={1--11},
  year={2019}
}

@inproceedings{wang2023dire,
  title={DIRE for Diffusion-Generated Image Detection},
  author={Wang, Zhendong and Bao, Jianmin and Zhou, Wengang and Wang, Weilun and He, Hao and Li, Houqiang},
  booktitle={Proceedings of the IEEE/CVF International Conference on Computer Vision (ICCV)},
  pages={22445--22455},
  year={2023}
}

@inproceedings{wang2020cnn,
  title={CNN-Generated Images Are Surprisingly Easy to Spot... For Now},
  author={Wang, Sheng-Yu and Wang, Oliver and Zhang, Richard and Owens, Andrew and Efros, Alexei A},
  booktitle={Proceedings of the IEEE/CVF Conference on Computer Vision and Pattern Recognition (CVPR)},
  pages={8695--8704},
  year={2020}
}

@inproceedings{qian2020thinking,
  title={Thinking in Frequency: Face Forgery Detection by Mining Frequency-aware Clues},
  author={Qian, Yuyang and Yin, Guojun and Sheng, Lu and Chen, Zenzhao and Shao, Jing},
  booktitle={European Conference on Computer Vision (ECCV)},
  pages={86--103},
  year={2020}
}

@inproceedings{liu2021spatial,
  title={Spatial-Phase Shallow Learning: Rethinking Face Forgery Detection in Frequency Domain},
  author={Liu, Hongguang and Li, Xiaodan and Zhou, Wenbo and Chen, Yuefeng and He, Yuan and Xue, Hui and Zhang, Weiming and Yu, Nenghai},
  booktitle={Proceedings of the IEEE/CVF Conference on Computer Vision and Pattern Recognition (CVPR)},
  pages={772--781},
  year={2021}
}

@inproceedings{cao2022end,
  title={End-to-End Reconstruction-Classification Learning for Face Forgery Detection},
  author={Cao, Junyi and Ma, Chao and Yao, Taiping and Chen, Shen and Ding, Shouhong and Yang, Xiaokang},
  booktitle={Proceedings of the IEEE/CVF Conference on Computer Vision and Pattern Recognition (CVPR)},
  pages={4113--4122},
  year={2022}
}

@inproceedings{ojha2023towards,
  title={Towards Universal Fake Image Detectors that Generalize Across Generative Models},
  author={Ojha, Utkarsh and Li, Yuheng and Lee, Yong Jae},
  booktitle={Proceedings of the IEEE/CVF Conference on Computer Vision and Pattern Recognition (CVPR)},
  pages={16014--16023},
  year={2023}
}

@inproceedings{li2020face,
  title={Face X-Ray for More General Face Forgery Detection},
  author={Li, Lingzhi and Bao, Jianmin and Zhang, Ting and Yang, Hao and Chen, Dong and Wen, Fang and Guo, Baining},
  booktitle={Proceedings of the IEEE/CVF Conference on Computer Vision and Pattern Recognition (CVPR)},
  pages={5001--5010},
  year={2020}
}

@inproceedings{tan2024npr,
  title={Rethinking Up-sampling Operations in {CNNs} for {AI}-Generated Image Detection},
  author={Tan, Chuangchuang and Zhao, Yao and Wei, Yunchao and Wang, Guanghua and Liu, Huan},
  booktitle={Proceedings of the IEEE/CVF Conference on Computer Vision and Pattern Recognition (CVPR)},
  pages={8974--8983},
  year={2024}
}

@inproceedings{chen2024drct,
  title={{DRCT}: Diffusion Reconstruction and Contrastive Tuning for {AI}-Generated Image Detection},
  author={Chen, Baoying and Zhou, Wengang and Wang, Weilun and Li, Houqiang},
  booktitle={International Conference on Learning Representations (ICLR)},
  year={2024}
}

@inproceedings{tan2025c2pclip,
  title={{C2P-CLIP}: Injecting Category Common Prompt in {CLIP} to Enhance Generalization in Deepfake Detection},
  author={Tan, Chuangchuang and Tao, Renshuai and Liu, Huan and Gu, Guanghua and Wu, Baoyuan and Zhao, Yao and Wei, Yunchao},
  booktitle={Proceedings of the AAAI Conference on Artificial Intelligence (AAAI)},
  year={2025}
}

@inproceedings{yang2025d3,
  title={{D$^3$}: Scaling Up Deepfake Detection by Learning from Discrepancy},
  author={Yang, Yongqi and Qian, Zhihao and Zhu, Ye and Russakovsky, Olga and Wu, Yu},
  booktitle={Proceedings of the IEEE/CVF Conference on Computer Vision and Pattern Recognition (CVPR)},
  year={2025}
}

@inproceedings{li2025rrbench,
  title={Bridging the Gap Between Ideal and Real-world Evaluation: Benchmarking {AI}-Generated Image Detection in Challenging Scenarios},
  author={Li, Chunxiao and Wang, Xiaoxiao and Li, Meiling and Miao, Boming and Sun, Peng and Zhang, Yunjian and Ji, Xiangyang and Zhu, Yao},
  booktitle={Proceedings of the IEEE/CVF Conference on Computer Vision and Pattern Recognition (CVPR)},
  year={2025}
}

@inproceedings{wang2025forensicsbench,
  title={Forensics-Bench: A Comprehensive Forgery Detection Benchmark Suite for Large Vision-Language Models},
  author={Wang, Jin and Lv, Chenghui and Li, Xian and Dong, Shichao and Li, Huadong and Yao, Kelu and Li, Chao and Shao, Wenqi and Luo, Ping},
  booktitle={Proceedings of the IEEE/CVF Conference on Computer Vision and Pattern Recognition (CVPR)},
  year={2025}
}

@inproceedings{guo2025m2f2det,
  title={Rethinking Vision-Language Model in Face Forensics: Multi-Modal Interpretable Forged Face Detector},
  author={Guo, Xiao and Song, Xiufeng and Zhang, Yue and Liu, Xiaohong and Liu, Xiaoming},
  booktitle={Proceedings of the IEEE/CVF Conference on Computer Vision and Pattern Recognition (CVPR)},
  year={2025}
}

@inproceedings{wang2023altfreezing,
  title={{AltFreezing} for More General Video Face Forgery Detection},
  author={Wang, Zhendong and Bao, Jianmin and Zhou, Wengang and Wang, Weilun and Li, Houqiang},
  booktitle={Proceedings of the IEEE/CVF Conference on Computer Vision and Pattern Recognition (CVPR)},
  pages={4129--4138},
  year={2023}
}

@inproceedings{yan2023ucf,
  title={{UCF}: Uncovering Common Features for Generalizable Deepfake Detection},
  author={Yan, Zhiyuan and Zhang, Yong and Fan, Xinchao and Wu, Baoyuan},
  booktitle={Proceedings of the IEEE/CVF International Conference on Computer Vision (ICCV)},
  pages={22412--22423},
  year={2023}
}

@article{bai2023qwen,
  title={Qwen-VL: A Versatile Vision-Language Model for Understanding, Localization, Text Reading, and Beyond},
  author={Bai, Jinze and Bai, Shuai and Yang, Shusheng and Wang, Shijie and Tan, Sinan and Wang, Peng and Lin, Junyang and Zhou, Chang and Zhou, Jingren},
  journal={arXiv preprint arXiv:2308.12966},
  year={2023}
}

@article{qwen2024qwen2vl,
  title={Qwen2-VL: To See the World More Clearly},
  author={Wang, Peng and Bai, Shuai and Tan, Sinan and Wang, Shijie and Fan, Zhihao and Bai, Jinze and Chen, Keqin and Liu, Xuejing and Wang, Jialin and Ge, Wenbin and others},
  journal={arXiv preprint arXiv:2409.12191},
  year={2024}
}

@article{liu2024visual,
  title={Visual Instruction Tuning},
  author={Liu, Haotian and Li, Chunyuan and Wu, Qingyang and Lee, Yong Jae},
  journal={Advances in Neural Information Processing Systems (NeurIPS)},
  volume={36},
  year={2023}
}

@inproceedings{xuan2025shieldhead,
  title={{ShieldHead}: Decoding-time Safeguard for Large Language Models},
  author={Xuan, Keqing and others},
  booktitle={Findings of the Association for Computational Linguistics: ACL 2025},
  pages={15582--15598},
  year={2025}
}

@article{glinerguard2026,
  title={{GLiNER} Guard: Unified Encoder Family for Production {LLM} Safety and Privacy},
  author={{GLiNER Team}},
  journal={arXiv preprint arXiv:2605.05277},
  year={2026}
}

@article{chen2016training,
  title={Training Deep Nets with Sublinear Memory Cost},
  author={Chen, Tianqi and Xu, Bing and Zhang, Chiyuan and Guestrin, Carlos},
  journal={arXiv preprint arXiv:1604.06174},
  year={2016}
}

@article{almeida2026jev,
  title={Jev: A Non-Autoregressive System One Decision Foundation Model for Structured Probabilistic Reasoning},
  author={Almeida, Diogo and {TypeSafe AI Team}},
  journal={TypeSafe AI Technical Report},
  year={2026}
}

@misc{laya2026,
  title={Laya: An Open-Source Non-Autoregressive Decision Foundation Model with Calibrated Typed Decision Heads},
  author={{ConvAI Innovations}},
  year={2026},
  howpublished={\url{https://huggingface.co/convaiinnovations/laya}},
  note={Open-weights release (322M, revision \texttt{main}), accessed Sept. 2026}
}

@inproceedings{wen2025fakeclue,
  title={Spot the Fake: Large Multimodal Model-Based Synthetic Image Detection with Artifact Explanation},
  author={Wen, Siwei and Ye, Junyan and Feng, Peilin and Kang, Hengrui and Wen, Zichen and Chen, Yize and Wu, Jiang and Wu, Wenjun and He, Conghui and Li, Weijia},
  booktitle={Advances in Neural Information Processing Systems (NeurIPS)},
  year={2025}
}

@article{singguard2026,
  title={SingGuard: An Efficient Decision Model for Large Language Model Guardrails},
  author={Tan, Wei and Ding, Liang and Fang, Meng and Liu, Yu and Shi, Shuming and Tao, Dacheng},
  journal={arXiv preprint arXiv:2606.22873},
  year={2026}
}

@misc{kwok2026clm,
  title={Contrastive Language Models: A System One Model for Fast and Generalizable Decision-Making},
  author={Kwok, Jacky and Kang, Hangoo and Suresh, Tarun and Saad-Falcon, Jon and Pavone, Marco and R{\'e}, Christopher and Mirhoseini, Azalia},
  year={2026},
  howpublished={Stanford University \& NVIDIA Research Technical Report},
  note={\url{https://contrastive-lm.notion.site} (\texttt{CLM-v0.1-8B})}
}

@article{zhu2026fakevlmr1,
  title={{FakeVLM-R1}: Internalizing Physical Laws via {CoT} for Synthetic Image Detection},
  author={Zhu, Leqi and Ye, Junyan and Lin, Kaiqing and Yan, Zhiyuan and others},
  journal={arXiv preprint arXiv:2605.30062},
  year={2026}
}

@article{lin2026deepvrm,
  title={Deep Residual Injection for Full-Spectrum Forensic Signal Perception in Multimodal Large Language Models},
  author={Lin, Kaiqing and Yan, Zhiyuan and Chen, Ruoxin and Zhang, Ke-Yue and others},
  journal={arXiv preprint arXiv:2606.15880},
  year={2026}
}

@article{wu2026foreagent,
  title={Perception, Verdict, and Evolution: Hindsight-Driven Self-Refining Forensics Agent for {AI}-Generated Image Detection},
  author={Wu, Yangjun and Yan, Keyu and Liu, Yu and Zhou, Jingren},
  journal={arXiv preprint arXiv:2606.26552},
  year={2026}
}

@article{yeh2026efr,
  title={Evidence-Grounded Forensic Reasoning for Detecting and Grounding Multi-Modal Media Manipulation},
  author={Yeh, Yichun and Li, Yiheng and Hu, Xiaobo and Lei, Zhen},
  journal={arXiv preprint arXiv:2608.08009},
  year={2026}
}

@article{li2026vigil,
  title={{VIGIL}: Part-Grounded Structured Reasoning for Generalizable Deepfake Detection},
  author={Li, Xinghan and Xu, Junhao and Chen, Jingjing},
  journal={arXiv preprint arXiv:2603.21526},
  year={2026}
}

@article{nguyen2025prpo,
  title={{PRPO}: Paragraph-level Policy Optimization for Vision-Language Deepfake Detection},
  author={Nguyen, Tuan and Khan, Naseem and Tran, Khang and Phan, NhatHai},
  journal={arXiv preprint arXiv:2509.26272},
  year={2025}
}

@article{choi2026ara,
  title={Anchor-Regularized Adaptation for Generalizable {AI}-Generated Image Detection with {DINOv3}},
  author={Choi, Hyeongjun and Lee, Juhun and Cozzolino, Davide and Verdoliva, Luisa and others},
  journal={arXiv preprint arXiv:2608.15196},
  year={2026}
}
\endgroup

\end{document}